\documentclass[11pt,a4paper]{article}
\usepackage[hyperref]{acl2020}
\usepackage{times}
\usepackage{latexsym}

\usepackage{microtype}
\usepackage{amsmath,amsfonts,amssymb}
\usepackage{booktabs}
\usepackage{tabularx}
\usepackage{url}

\usepackage{float}
\usepackage{graphicx}

\aclfinalcopy 

\title{Windowing Models for Abstractive Summarization of Long Texts}

\author{Leon Sch\"{u}ller\textsuperscript{1,2}, Florian Wilhelm\textsuperscript{2}, Nico Kreiling\textsuperscript{2} and Goran Glava\v{s}\textsuperscript{1} \vspace{0.2em} \\
  \textsuperscript{1} Data and Web Science Group,  University of Mannheim \\
  \textsuperscript{2} inovex GmbH \vspace{0.2em} \\
  {\tt leon.schueller92@googlemail.com} \\
  {\tt goran@informatik.uni-mannheim.de} \\
}
\date{}

\begin{document}
\maketitle
\begin{abstract}
  Neural summarization models suffer from the fixed-size input limitation: if text length surpasses the model's maximal number of input tokens, some document content (possibly summary-relevant) gets truncated Independently summarizing windows of maximal input size disallows for information flow between windows and leads to incoherent summaries. We propose windowing models for neural abstractive summarization of (arbitrarily) long texts. We extend the sequence-to-sequence model augmented with pointer generator network by (1) allowing the encoder to slide over different windows of the input document and (2) sharing the decoder and retaining its state across different input windows. We explore two windowing variants: Static Windowing precomputes the number of tokens the decoder should generate from each window (based on training corpus statistics); in Dynamic Windowing the decoder learns to emit a token that signals encoder's shift to the next input window. Empirical results render our models effective in their intended use-case: summarizing long texts with relevant content not bound to the very document beginning.
  \end{abstract}

\section{Background and Motivation}


While extractive summarization selects and copies the most relevant source phrases and sentences to the summary, abstractive summarization (AS) aims to capture the source meaning and generate summaries not necessarily containing portions of the source texts \cite{DBLP:journals/ftir/NenkovaM11}, holding promise of producing summaries more like human created ones. 
State-of-the-art neural AS models \cite{NallapatiXZ16,see2017get,Paulus18,P17-1108,makino-etal-2019-global,you-etal-2019-improving} extend a standard sequence-to-sequence (Seq2Seq) architecture, using either recurrent (RNN) \cite{Bahdanau15} or Transformer-based \cite{vaswani2017attention} encoder and decoder components. 
\newcite{see2017get} extend the standard Seq2Seq model with a pointer-generator network (PG-Net), providing the model with extractive capabilities, i.e., allowing it to choose between generating a token and copying source text tokens. \newcite{P17-1108} propose a hierarchical model that introduces an additional graph-based attention mechanism which serves to model interactions between encoded sentence representations. \newcite{Paulus18} incorporate a reward expectation based on reinforcement learning into a mixed training objective to steer the model towards predicting globally meaningful sequences. 

With respect long-document summarization, \newcite{celikyilmaz2018deep} distribute the encoding task to multiple collaborating encoder agents, whereas \newcite{cohan2018discourse} propose a hierarchical encoder that captures the document's discourse structure, and an attentive discourse-aware
decoder that generates the summary. The latter requires a predefined discourse structure and is designed for domain-specific texts (e.g., scientific publications). Despite multiple encoders operating on different document segments, these models still limit the maximal document length at inference. 


In this work, we address a prominent limitation of neural AS models: they cannot summarize texts longer than the maximal input length $T_x$ set during model training. At inference, documents longer than $T_x$ tokens are truncated, which renders the (potentially summary-relevant) truncated content inaccessible to the model.
We propose novel AS models based on windowing of source text: we sequentially shift encoder's attention over different windows of source text. The decoder is shared across windows, thereby preserving semantic information from a previous window when decoding the next. We investigate two windowing strategies: (1) Static Windowing Model (SWM) precomputes, based on the training corpus statistics, the number of tokens the decoder is to generate from each source window; (2) for the Dynamic Windowing Model (DWM), we first heuristically, based on semantic similarity between source text and summary sentences, inject special \textit{window-shift} tokens into the training reference summaries and then let the decoder learn to emit window-shift tokens during generation. Signaling the window shift by generating a special token, conceptually allows the DWM model to summarize arbitrarily long texts during inference. Evaluation on the WikiHow corpus \cite{wikihow} of long texts with more even distribution of summary-relevant content renders our windowing models effective.


\section{Windowing AS Models}
\label{sec:windowing}

\begin{figure}
    \centering
    \includegraphics[scale=0.6]{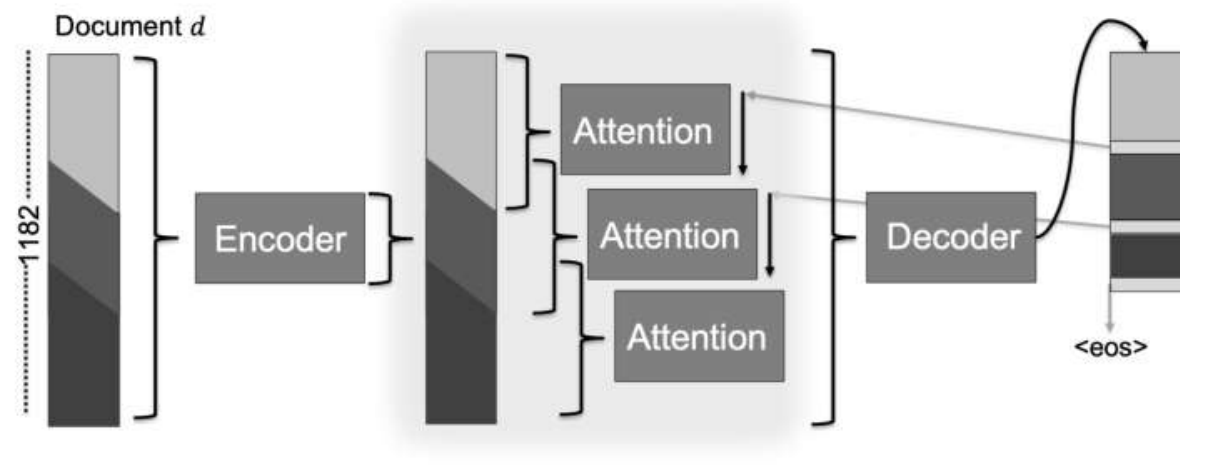}
    \caption{High-level illustration of the windowing model for long document summarization.}
    \label{fig:model}
    \vspace{-1em}
\end{figure}

Figure \ref{fig:model} contains the high-level depiction of the windowing AS model. We start from the attention-based Seq2Seq model with recurrent components \cite{Bahdanau15},\footnote{We also experimented with Transformer \cite{vaswani2017attention} encoder/decoder, but obtained weaker performance.} which maps the input sequence $x_1,...,x_{T_x}$ into an output sequence $y_1,...,y_{T_y}$. 
%
A bidirectional LSTM (Bi-LSTM) encoder produces contextualized representations $h_j = [\overrightarrow{h}_j ; \overleftarrow{h}_j]$ for each input token. 
Decoder's state is initialized with the concatenation of the end states of encoder's LSTMs ($s_0 = [\overrightarrow{h}_{T_x} ; \overleftarrow{h}_{1}]$). 
%
We apply an attention mechanism similar to \newcite{luong15effective}. However, instead of learning a local attention span around each source text position -- which would limit the model to a fixed-size input during training -- we attend over a window of $T_w$ tokens and sequentially slide the window over the long text. This way the decoder learns to model transitions between content windows, allowing to summarize arbitrarily long documents at inference.  

Window size $T_w$ and a stride step $ss$, divide the source text  ($T_x$ tokens) into overlapping windows.\footnote{We pad the last window(s), if shorter than $T_w$ tokens.} We use the same decoder, retaining its state, across all input windows. 
Sharing a decoder across input windows allows the flow of semantic information between adjacent windows and holds promise of retaining summary coherence.
%
At each decoding step $t$, we attend over the window representations, using the decoder's hidden state $s_t$ as the attention query, and obtain the conditioned window encoding $c_t$ (for the decoding step $t$):
$c_t = \sum_{j \in T_w} \alpha_{t,j} h_j$, with attention weight $\alpha_{t,j}$ computed as the softmax-normalized value of the dot-product $s_t^\top h_j$ between the encoded token $h_j$ and the decoder's state $s_t$. Decoder outputs the embedding $l_t$ via feed-forward projection of the concatenation of the attended input representation $c_t$ and its own hidden state $s_t$: $l_t = W_l\, \mathrm{tanh}([c_t; s_t]) + b_l$, with $W_l \in \mathbb{R}^{d\ \times\ 2d}, b_l \in \mathbb{R}^d$ as parameters. The output probability distribution $P_V$ (over training vocabulary $V$) is then simply computed by applying the softmax function on the vector of dot-product values computed between $l_t$ and each of the (pretrained) word embeddings.

We augment the base Seq2Seq model with the pointer-generator network (PG-Net), as in \cite{see2017get}, allowing the decoder to choose, in each step, between generating a token from the training vocabulary and copying a token from the source document. Generation probability is based on the context vector $c_t$, decoder's hidden state $s_t$, and decoder's input $x_t$: 
\begin{equation}
p_{\mathit{gen}} = \sigma(w^\top_{c} c_t + w^\top_s s_t + w^\top_x x_t + b_{\mathit{ptr}})    
\end{equation}

\noindent with $w_c$, $w_s$ $\in \mathbb{R}^d$, $w_x \in \mathbb{R}^{d_{emb}}, b_{\mathit{ptr}} \in \mathbb{R}$ as parameters. The output probability for a word $x$ from the extended vocabulary $\hat{V}$ (union of $V$ and source text words) interpolates between generation and copying distributions:  
\begin{equation}
P_{\hat{V}}(x) = p_{\mathit{gen}} \cdot P_\mathit{V}(x) + (1 - p_{\mathit{gen}})\sum_{j:x_j = x}{\alpha_{t,j}}
\end{equation}


This specifies the PG-Net-augmented Seq2Seq AS model that operates on a window ($T_w$ tokens). We next need to specify when to transition from one window of source text to another.

\subsection{Static Windowing Model}

The Static Windowing Model precomputes the number of tokens the decoder needs to generate for each input window. 
Let $\{w_1, w_2, \dots, w_N\}$ be the equally-sized source windows (determined with $T_w$ and $ss$). We use the following function to determine the importance (weight) for each window: $e_s(w_i) = \mathrm{exp}(-k (1 + i\cdot d^i))$, with $k$ and $d$ as parameters defining the shape of the summary distribution over windows.\footnote{For example, with $d=1.2$ and $k=0.8$, the early windows will receive larger weights than the later windows.} The unnormalized weights $e_s(w_i)$ are converted into probabilities using the softmax function.  
%
We next compute the expected summary length for a given document, based on the document length and training corpus statistics. Let $D$ be the set of documents and $S$ the set of their respective reference summaries in the training corpus. We compute the expected summary length for a new document $d$ as:
\begin{equation}
\mathbb{E}(|s|)_{d} = \mathit{majority}(|S|) \cdot \frac{|d|}{\mathit{majority}(|D|)}    
\end{equation}

\noindent where $\mathit{majority}(|D|)$ is the length that covers 90\% of training documents (i.e., 90\% of $d \in D$ are at most $\mathit{majority}(|D|)$) and $\mathrm{Majority}(|S|)$ is the length that covers 90\% of reference summaries from $S$. The number of tokens the decoder is to generate for a window $w_i$ is now simply a product of $\mathbb{E}(|s|)_{d}$ and the normalized weight $e_s(w_i)$. 



\subsection{Dynamic Windowing Model}
SWM still relies on the document (and summary) lengths of the training corpus, and the number of summary tokens decoded for a window does not it's content.  
Dynamic Windowing Model (DWM) aims to be more flexible, by allowing the decoder to dynamically signal, via a special token, the saturation of the current window and shift to the next. Because (1) the decoder needs to learn to emit this window-shift token ($\rightarrow$), and (2) we still want an end-to-end trainable AS model, we need to somehow inject window-shift tokens ($\rightarrow$) into reference summaries of the training corpus. We achieve this heuristically, by computing semantic similarity scores between source text sentences and reference summary sentences. For simplicity, we obtain the sentence embedding as a sum of its respective word embeddings and compute the cosine similarity between sentence embeddings.\footnote{We acknowledge that this is a rudimentary method for computing semantic similarity between sentences. We intend to experiment with more advanced sentence embedding models and more accurate sentence similarity measures \cite[\textit{inter alia}]{kusner2015word,conneau2017supervised,devlin2019bert,zhelezniak2019don} in subsequent work. 
}  
%

For every reference summary sentence, we identify the most similar source document sentence and determine its respective window.\footnote{Depending on $T_w$ and $ss$, a sentence be in more than one window. In such cases, we map the sentence to the last containing window.} This way we map each reference summary sentence to one source window. The order of windows assigned to summary sentences is, however, not necessarily sequential (e.g., $[1,3,2,4,3]$ for some reference summary with five sentences). Since our model allows only sequential window shifts, we first make the window order sequential by replacing sequence-breaking windows with accumulated maximums (e.g., $[1,3,2,4,3]$ becomes $[1,3,3,4,4]$). We then inject window-shift tokens ($\rightarrow$) between summary sentences with different assigned source windows (e.g., for the window assignment $[1,3,3,4,4]$ we inject $\rightarrow$ $\rightarrow$ between the first and second summary sentence and $\rightarrow$ between the third and fourth sentence).  
%
%
During inference, the input window is shifted whenever the decoder outputs the $\rightarrow$ token. 

\section{Evaluation}

\paragraph{Data.} We evaluate our windowing models on two benchmark datasets: (1) CNN/Dailymail news corpus, created by \cite{NallapatiXZ16} from the question answering dataset of \newcite{hermann2015teaching} and (2) WikiHow corpus \cite{wikihow}. News place the most relevant information at the beginning (the so-called lead-and-body principle): the standard models that truncate long documents are thus likely to perform well in the CNN/Dailymail evaluation. The WikiHow dataset does not have such a construction bias -- summary-relevant information is more evenly distributed across the texts.

\paragraph{Experimental Setup.} 
We use the negative log likelihood objective and optimize the models by maximizing the ROUGE-L performance on development sets.
We use a batch-level beam search decoder with beam size $B=3$. Unlike standard beam search, $B$ does not decrease when the end-of-summary token (\textless eos\textgreater) is predicted. Longer yet incomplete partial hypotheses can thus take over completed beams whenever they prevail in terms of length-normalized log probability. We set the hidden state sizes for both encoder's LSTMs and decoder's LSTM to $256$.
We employ the Adam optimizer \cite{DBLP:journals/corr/KingmaB14} ($\beta_1$=0.9, $\beta_2$=0.999, and $\epsilon$=1e-8). 
For word representations, we use pretrained 300-dim.~fastText embeddings (50,000 most frequent words)\footnote{\url{https://tinyurl.com/y3y69h3z}} 

\paragraph{Baselines.} We compare different variants of SWM and DWM against the standard PG-Net Seq2Seq model (\textsc{Stan}) with the fixed-size input \cite{see2017get}, as well as against the commonly employed \textsc{Lead-3} baseline, which simply copies the first three document sentences to the summary.

\paragraph{Results and Discussion.}

\setlength{\tabcolsep}{15pt}
\begin{table}[t]
\centering
\def\arraystretch{0.87}
\vspace{-0.0em}
{\footnotesize
\begin{tabularx}{\linewidth}{l ccc}
\toprule
Model & R-1 & R-2 & R-L \\ \midrule
\textsc{Lead-3} & 39.89 & 17.22 & 36.08 \\ 
\textsc{Stan} & 37.85 & 16.48 & 34.95 \\ \midrule
SWM & 37.11 & 16.01 & 34.37 \\
DWM & 36.02 & 15.67 & 33.28 \\
\bottomrule
\end{tabularx}
}
\vspace{-0.0mm}
\caption{Results on the CNN/Dailymail test set: summaries of $T_y=125$ tokens; \textsc{Stan} trained with fixed-size input of $T_x=400$ tokens; SWM ($d=1.2$, $k=0.8$) \& DWM trained on $T_x=1160$ tokens, with windows of $T_w = 400$ tokens (stride $ss=380$).}
\vspace{-0.0mm}
\label{tbl:cnndm}
\end{table}

Table \ref{tbl:cnndm} contains the results on the CNN/Dailymail dataset. Unsurprisingly, the simple \textsc{Lead-3} baseline outperforms \textit{Stan} and both our static and dynamic windowing models. This is because in CNN/Dailymail documents almost all of the summary-relevant content is found at the very beginning of the document. The ability to process all windows does not benefit to SWM and DWM in this setting as there is virtually no summary-relevant content in later windows.  


In Table \ref{tbl:wikihow} we display the results on the WikiHow dataset, which is bound to be more appropriate for the windowing models, because of the more even distribution of the summary-relevant content across the source documents.
\setlength{\tabcolsep}{7pt}
\begin{table}[t]
\centering
\def\arraystretch{0.87}
\vspace{-0.0em}
{\footnotesize
\begin{tabularx}{\linewidth}{lll ccc}
\toprule
Model & $T_x$ & $T_w/ss$ & R-1 & R-2 & R-L \\ \midrule
\textsc{Lead-3} & -- & -- & 24.24 & 5.31 & 21.86 \\ \midrule
\textsc{Stan} & \textbf{200} & - & 22.84 & 7.89 & 22.38 \\
DWM & 740 & \textbf{200}/180 & 26.15 & 8.63 & 25.48 \\ \midrule
\textsc{Stan} & \textbf{400} & - & 27.54 & 9.59 & 26.85 \\
SWM & 780 & \textbf{400}/380 & 28.25 & 9.71 & 27.55 \\
DWM & 780 & \textbf{400}/380 & 27.23 & 9.51 & 26.49 \\
\bottomrule
\end{tabularx}
}
\caption{Results on the WikiHow dataset ($T_y=125$, $d=0$ for SWM).}
\vspace{-1em}
\label{tbl:wikihow}
\end{table}
On the WikiHow dataset, the windowing models -- SWM and DWM -- generally have an edge over the standard PG-Net Seq2Seq model (\textsc{Stan}) when the fixed-size input for \textsc{Stan} matches the windows size of the windowing models. For a larger input size $T_x=400$, \textsc{Stan} performs comparably to DWM with the same window size $T_w=400$. Notably, the DWM has the advantage of being able to process longer overall input.
Lowering $T_x$ for \textsc{Stan} to $200$ and comparing it against SWM/DWM with windows of the same size $T_w=200$, we see that the windowing models clearly prevail. This renders our windowing models as a more approriate solution for summarization of documents for which the following two properties hold: (1) the document length massively surpasses the maximal number of tokens we can feed to the fixed-input-size model and (2) summary-relevant information is present all across the document, and not just at its beginning.

\begin{figure}
  \includegraphics[width=\linewidth]{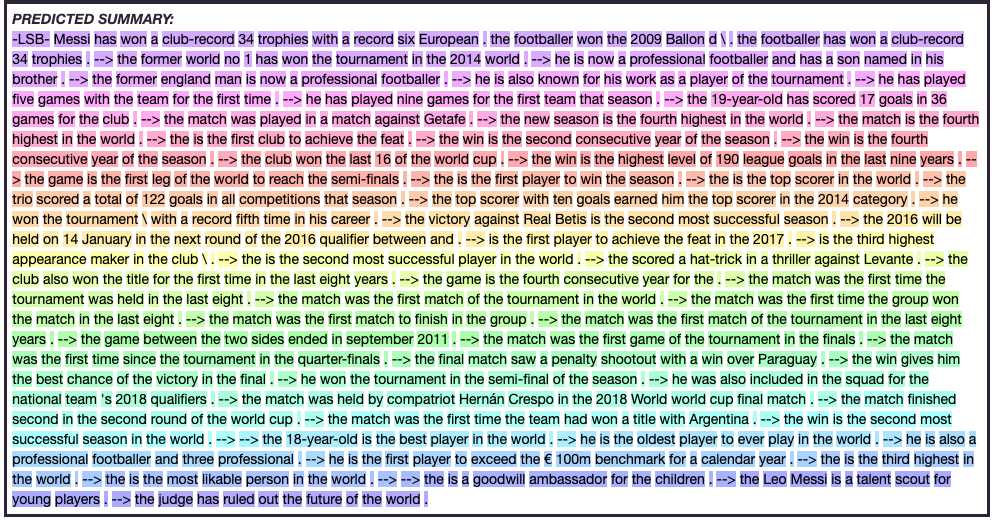}
  \caption{Summary for the Wikipedia page ``Lionel Messi'' ($13.607$ tokens) produced by DWM trained on CNN/Dailymail ($T_x=1.160$ tokens). Colors correspond to different source text windows over which the decoder attended during generation.}
  \label{fig:standard}
\end{figure}

While SWM seems to outperform DWM, in practice SWM cannot really summarize arbitrarily long texts at inference. Despite transitioning across content windows, SWM adapts to the summary lengths seen in the training corpus and generates the\, \textless eos\textgreater\, token too early during inference on the long texts. 
In contrast, by learning to emit window transitions, the Dynamic Windowing Model can truly generate summaries for arbitrarily long texts at inference time, regardless of the observed lengths of training document and their respective reference summaries. Figure \ref{fig:standard} depicts the summary of a very long document ($13.607$ tokens), produced by a DWS model trained on an order of magnitude shorter documents ($T_x=1.160$ tokens).

\section{Conclusion}

Neural summarization models fix the length of the source texts in training (e.g., based on the average source document length in the training set), forcing documents longer than this threshold to be truncated at inference. In this work, we proposed windowing summarization models, which allow to process arbitrarily long documents at inference, taking into account full source text. Our models are effective in summarizing long texts with evenly distributed summary-relevant content.

\bibliographystyle{acl_natbib}
\bibliography{references}

\begin{thebibliography}{19}
\expandafter\ifx\csname natexlab\endcsname\relax\def\natexlab#1{#1}\fi

\bibitem[{Bahdanau et~al.(2015)Bahdanau, Cho, and Bengio}]{Bahdanau15}
Dzmitry Bahdanau, Kyunghyun Cho, and Yoshua Bengio. 2015.
\newblock \href {http://arxiv.org/abs/1409.0473} {Neural machine translation by
  jointly learning to align and translate}.
\newblock In \emph{Proceedings of ICLR}.

\bibitem[{Celikyilmaz et~al.(2018)Celikyilmaz, Bosselut, He, and
  Choi}]{celikyilmaz2018deep}
Asli Celikyilmaz, Antoine Bosselut, Xiaodong He, and Yejin Choi. 2018.
\newblock Deep communicating agents for abstractive summarization.
\newblock In \emph{Proceedings of the 2018 Conference of the North American
  Chapter of the Association for Computational Linguistics: Human Language
  Technologies, Volume 1 (Long Papers)}, pages 1662--1675.

\bibitem[{Cohan et~al.(2018)Cohan, Dernoncourt, Kim, Bui, Kim, Chang, and
  Goharian}]{cohan2018discourse}
Arman Cohan, Franck Dernoncourt, Doo~Soon Kim, Trung Bui, Seokhwan Kim, Walter
  Chang, and Nazli Goharian. 2018.
\newblock A discourse-aware attention model for abstractive summarization of
  long documents.
\newblock In \emph{Proceedings of the 2018 Conference of the North American
  Chapter of the Association for Computational Linguistics: Human Language
  Technologies, Volume 2 (Short Papers)}, pages 615--621.

\bibitem[{Conneau et~al.(2017)Conneau, Kiela, Schwenk, Barrault, and
  Bordes}]{conneau2017supervised}
Alexis Conneau, Douwe Kiela, Holger Schwenk, Lo{\"\i}c Barrault, and Antoine
  Bordes. 2017.
\newblock Supervised learning of universal sentence representations from
  natural language inference data.
\newblock In \emph{Proceedings of the 2017 Conference on Empirical Methods in
  Natural Language Processing}, pages 670--680.

\bibitem[{Devlin et~al.(2019)Devlin, Chang, Lee, and
  Toutanova}]{devlin2019bert}
Jacob Devlin, Ming-Wei Chang, Kenton Lee, and Kristina Toutanova. 2019.
\newblock Bert: Pre-training of deep bidirectional transformers for language
  understanding.
\newblock In \emph{Proceedings of the 2019 Conference of the North American
  Chapter of the Association for Computational Linguistics: Human Language
  Technologies, Volume 1 (Long and Short Papers)}, pages 4171--4186.

\bibitem[{Hermann et~al.(2015)Hermann, Kocisky, Grefenstette, Espeholt, Kay,
  Suleyman, and Blunsom}]{hermann2015teaching}
Karl~Moritz Hermann, Tomas Kocisky, Edward Grefenstette, Lasse Espeholt, Will
  Kay, Mustafa Suleyman, and Phil Blunsom. 2015.
\newblock Teaching machines to read and comprehend.
\newblock In \emph{Advances in neural information processing systems}, pages
  1693--1701.

\bibitem[{Kingma and Ba(2015)}]{DBLP:journals/corr/KingmaB14}
Diederik~P. Kingma and Jimmy Ba. 2015.
\newblock Adam: {A} method for stochastic optimization.
\newblock In \emph{{ICLR}}.

\bibitem[{Koupaee and Wang(2018)}]{wikihow}
Mahnaz Koupaee and William~Yang Wang. 2018.
\newblock \href {http://arxiv.org/abs/1810.09305} {Wikihow: {A} large scale
  text summarization dataset}.
\newblock \emph{CoRR}, abs/1810.09305.

\bibitem[{Kusner et~al.(2015)Kusner, Sun, Kolkin, and
  Weinberger}]{kusner2015word}
Matt Kusner, Yu~Sun, Nicholas Kolkin, and Kilian Weinberger. 2015.
\newblock From word embeddings to document distances.
\newblock In \emph{International conference on machine learning}, pages
  957--966.

\bibitem[{Luong et~al.(2015)Luong, Pham, and Manning}]{luong15effective}
Minh{-}Thang Luong, Hieu Pham, and Christopher~D. Manning. 2015.
\newblock \href {http://arxiv.org/abs/1508.04025} {Effective approaches to
  attention-based neural machine translation}.
\newblock In \emph{Proceedings of EMNLP}, pages 1412--1421.

\bibitem[{Makino et~al.(2019)Makino, Iwakura, Takamura, and
  Okumura}]{makino-etal-2019-global}
Takuya Makino, Tomoya Iwakura, Hiroya Takamura, and Manabu Okumura. 2019.
\newblock \href {https://www.aclweb.org/anthology/P19-1099} {Global
  optimization under length constraint for neural text summarization}.
\newblock In \emph{Proceedings of the 57th Annual Meeting of the Association
  for Computational Linguistics}, pages 1039--1048.

\bibitem[{Nallapati et~al.(2016)Nallapati, Xiang, and Zhou}]{NallapatiXZ16}
Ramesh Nallapati, Bing Xiang, and Bowen Zhou. 2016.
\newblock \href {http://arxiv.org/abs/1602.06023} {Sequence-to-sequence rnns
  for text summarization}.
\newblock In \emph{Proceedings of ICLR: Workshop Track}.

\bibitem[{Nenkova and McKeown(2011)}]{DBLP:journals/ftir/NenkovaM11}
Ani Nenkova and Kathleen~R. McKeown. 2011.
\newblock Automatic summarization.
\newblock \emph{Foundations and Trends in Information Retrieval},
  5(2-3):103--233.

\bibitem[{Paulus et~al.(2018)Paulus, Xiong, and Socher}]{Paulus18}
Romain Paulus, Caiming Xiong, and Richard Socher. 2018.
\newblock \href {http://arxiv.org/abs/1705.04304} {A deep reinforced model for
  abstractive summarization}.
\newblock In \emph{Proceedings of ICLR}.

\bibitem[{See et~al.(2017)See, Liu, and Manning}]{see2017get}
Abigail See, Peter~J Liu, and Christopher~D Manning. 2017.
\newblock Get to the point: Summarization with pointer-generator networks.
\newblock In \emph{Proceedings of the 55th Annual Meeting of the Association
  for Computational Linguistics (Volume 1: Long Papers)}, pages 1073--1083.

\bibitem[{Tan et~al.(2017)Tan, Wan, and Xiao}]{P17-1108}
Jiwei Tan, Xiaojun Wan, and Jianguo Xiao. 2017.
\newblock \href {https://doi.org/10.18653/v1/P17-1108} {Abstractive document
  summarization with a graph-based attentional neural model}.
\newblock In \emph{Proceedings of the 55th Annual Meeting of the Association
  for Computational Linguistics (Volume 1: Long Papers)}, pages 1171--1181.
  Association for Computational Linguistics.

\bibitem[{Vaswani et~al.(2017)Vaswani, Shazeer, Parmar, Uszkoreit, Jones,
  Gomez, Kaiser, and Polosukhin}]{vaswani2017attention}
Ashish Vaswani, Noam Shazeer, Niki Parmar, Jakob Uszkoreit, Llion Jones,
  Aidan~N Gomez, Lukasz Kaiser, and Illia Polosukhin. 2017.
\newblock Attention is all you need.
\newblock In \emph{Proceedings of NeurIPS}.

\bibitem[{You et~al.(2019)You, Jia, Liu, and Yang}]{you-etal-2019-improving}
Yongjian You, Weijia Jia, Tianyi Liu, and Wenmian Yang. 2019.
\newblock \href {https://www.aclweb.org/anthology/P19-1205} {Improving
  abstractive document summarization with salient information modeling}.
\newblock In \emph{Proceedings of the 57th Annual Meeting of the Association
  for Computational Linguistics}, pages 2132--2141.

\bibitem[{Zhelezniak et~al.(2019)Zhelezniak, Savkov, Shen, Moramarco, Flann,
  and Hammerla}]{zhelezniak2019don}
Vitalii Zhelezniak, Aleksandar Savkov, April Shen, Francesco Moramarco, Jack
  Flann, and Nils~Y Hammerla. 2019.
\newblock Don't settle for average, go for the max: Fuzzy sets and max-pooled
  word vectors.
\newblock In \emph{Proceedings of ICLR}.

\end{thebibliography}

\end{document}